\documentclass{article} %
\usepackage{iclr2021_conference,times}

\usepackage{amsmath,amsfonts,bm}

\def\eqref#1{equation~\ref{#1}}

\def\1{\bm{1}}

\DeclareMathAlphabet{\mathsfit}{\encodingdefault}{\sfdefault}{m}{sl}
\SetMathAlphabet{\mathsfit}{bold}{\encodingdefault}{\sfdefault}{bx}{n}

\usepackage{hyperref}
\usepackage{url}
\usepackage{enumitem}
\usepackage{booktabs}
\usepackage{amsthm}
\usepackage{mathrsfs}
\usepackage{multirow}
\usepackage{graphicx}
\usepackage{wrapfig}
\usepackage{algorithmic,algorithm}

\usepackage{colortbl}

\definecolor{mygray}{RGB}{230,230,250}
\definecolor{mygray2}{RGB}{176,196,222}

\title{Pathological Visual Question Answering}

\author{Xuehai He\thanks{equal contribution}\\
University of California San Diego\\
\texttt{x5he@ucsd.edu} \\
\And
Zhuo Cai\footnotemark[1] \\
Tsinghua University \\
\texttt{caiz17@mails.tsinghua.edu.cn} 
\And
Wenlan Wei \\
Wuhan University \\
\texttt{wwl999@whu.edu.cn} 
\And
Yichen Zhang \\
University of California San Diego \\
\texttt{yiz037@eng.ucsd.edu\quad\quad\quad\quad\quad\;\,} 
\And
Luntian Mou\\
Beijing University of Technology \\
\texttt{ltmou@pku.edu.cn } 
\And
Eric Xing\\
Carnegie Mellon University \\
\texttt{epxing@cs.cmu.edu\quad\quad\quad\quad\quad\quad\,\,\,\,} 
\And
Pengtao Xie \\
University of California San Diego \\
\texttt{pengtaoxie2008@gmail.com}
}

\iclrfinalcopy %
\begin{document}

\maketitle

\begin{abstract}
Is it possible to develop an ``AI Pathologist" to pass the board-certified examination of the American Board of Pathology (ABP)? To build such a system, three challenges need to be addressed. First, we need to create a visual question answering (VQA) dataset where  the AI agent is presented with a pathology image together with a question  and is asked to give the correct answer. 
Due to privacy concerns, pathology images are usually not publicly available. Besides, only well-trained  pathologists  can understand  pathology  images, but they barely have time to help create datasets for AI research. The second challenge is: due to the fact that it is difficult to hire highly experienced pathologists to create pathology visual questions and answers, the resulting pathology VQA dataset may contain errors such as some questions may not be relevant to the image or the answers are not given correctly. Training pathology VQA models using these noisy or even erroneous data will lead to problematic models that cannot generalize well on unseen images. The third challenge is: the medical concepts and knowledge covered in pathology question-answer (QA) pairs are very diverse while the number of QA pairs available for modeling training is limited. How  to learn effective representations of diverse medical concepts based on limited data is technically demanding. 
In this paper, we aim to address these three challenges. To our best knowledge, our work represents the first one addressing the pathology VQA problem. To deal with the issue that a publicly available pathology VQA dataset is lacking, we create PathVQA, a VQA dataset with 32,795  questions asked from 4,998 pathology images. The questions in PathVQA are similar to those in the ABP tests.  
To our best knowledge, this is the first dataset for pathology VQA. 
To address the second challenge, we propose a learning-by-ignoring approach which automatically identifies training examples 
that have bad-quality and remove them from the training dataset. 
To address the third challenge, we propose to use cross-modal self-supervised learning to learn powerful visual and  textual  representations jointly.
We perform experiments on our created PathVQA dataset and the results demonstrate the effectiveness of our proposed learning-by-ignoring method and cross-modal self-supervised learning methods. 
\end{abstract}
\vspace{-0.5cm}
\section{Introduction}
\vspace{-0.3cm}
Pathology studies the causes and effects of diseases or injuries. It underpins every aspect of patient care, such as diagnostic testing, providing treatment advice, preventing diseases using cutting-edge genetic technologies, to name a few. Medical professionals practicing pathology are called pathologists, who examine bodies and body tissues. To become a board-certificated pathologist in the US, a medical professional needs to pass a certification examination organized by the American Board of Pathology (ABP), which is a very challenging task. We are interested in asking: whether an artificial intelligence (AI) system can be developed to pass the ABP examination? It is an important step towards achieving AI-aided clinical decision support and clinical education. Among the ABP test questions, one major type is to understand the pathology images. Given a pathology image and a question, the examinees are asked to select a correct answer. Such a problem is called visual question answering (VQA)~\citep{vqa} in the AI community. VQA is an interdisciplinary research problem that has drawn extensive attention recently. Given an image (e.g., an image showing a dog is chasing a ball) and a question  asked about the visual content of the image (e.g., ``what is the dog chasing?"), VQA aims to develop AI algorithms to infer the correct answer (e.g., ``ball"). VQA requires a deep comprehension of both images and textual questions, as well as the relationship between visual objects and textual entities, which is technically very demanding.

To train an AI system to  perform VQA on pathology images and pass the ABP test, we first need to collect a dataset containing questions similar to those in the ABP test. ABP provides some sample questions, but they are too few to be useful for training data-driven models. Some commercial institutes provide a larger number of practice  questions, but they are very expensive to buy and they cannot be shared with the public due to copyright issues. One possible way to create pathology VQA dataset is to leverage crowdsourcing, which is used successfully for creating  general domain VQA datasets~\citep{first,vqa,cocoQA,clevr,goyal}. 
However, it is much more challenging to build medical VQA datasets than general domain VQA datasets via crowdsourcing. First, medical images such as pathology images are highly domain-specific, which can only be  interpreted by  well-educated medical professionals. It is very difficult and expensive to hire medical professionals to help create medical VQA datasets.
Second, to create a VQA dataset, one first needs to collect an image dataset. While images in the general domain are pervasive, medical images are very difficult to obtain due to privacy concerns.

To address these challenges, we resort to pathology textbooks, especially those that are freely accessible online, as well as online digital libraries. These textbooks contain a lot of pathology images, covering the entire domain of pathology. Each image has a caption describing pathological findings in the image. The caption is carefully worded and clinically precise. We extract images and captions from the textbooks and online digital libraries. Given these images, question-answer pairs are created based on image captions. These QA pairs are verified by medical professionals to ensure clinical meaningfulness and correctness. In the end, we create a pathology VQA dataset called PathVQA, which contains  32,795 questions asked from 4,998 pathology images.  To our best knowledge, this is the first dataset for pathology VQA. 

Given the pathology VQA dataset, the next step is to develop a pathology VQA system, which is also very challenging, due to the following reasons. First, while we have tried our best to ensure the clinical correctness of the PathVQA dataset, it may still contain noises and errors that can only be identified by very experienced pathologists who unfortunately do not have time to do so for all the data examples in PathVQA. To address this problem, we propose a learning-by-ignoring method which can automatically identify bad-quality data (errors, noises, outliers, etc.) and remove them from the training set. 
The learning-by-ignoring strategy analyzes the collection of training examples holistically and determines which ones should be ignored. The likelihood of ignoring each training example is learned by maximizing the performance on the validation set in a bi-level optimization framework. 
The second challenge is: the medical concepts involved in PathVQA are very diverse while the number of question-answer pairs available for training is limited. 
Learning effective representations of these diverse medical concepts using limited data is technically difficult. Poorly learned  representations lead to inferior VQA performance.
To address the second challenge,
we propose cross-modal self-supervised learning approaches to pretrain the representation learning modules in VQA models for obtaining effective visual and textual embeddings.
Self-supervised learning (SSL)~\citep{ssl1, ssl2, ssl3} is an unsupervised learning approach which creates auxiliary tasks on input data without using human-provided labels and learns data representations by solving these auxiliary tasks. We create two types of cross-modal SSL tasks: 1) given an image and a question, judge whether this question is asked from this image; 2) given an image and an answer, judge whether this answer is relevant to this image. We also conduct a single-modal SSL on question-answer pairs: we pretrain the text encoder by predicting answers only based on the questions without considering the input images. 
Experiments on our developed PathVQA dataset demonstrates the effectiveness of our proposed methods.

The major contributions of this paper are as follows:
\begin{itemize}[leftmargin=*]
\setlength\itemsep{0em}
\vspace{-0.3cm}
\item
We create a pathology visual question answering dataset -- PathVQA, to foster the research of medical VQA. To our best knowledge, this is the first dataset for pathology VQA.
\item We propose a learning-by-ignoring approach which automatically identifies problematic training examples and removes them from the training set. Our method performs data ignoring by maximizing the validation performance end-to-end.
\item We propose  cross-modal self-supervised learning (SSL) approaches to learn better image encoders and text encoders in VQA models.
Three SSL strategies are studied, including 1) predicting whether an image and a question match, 2) predicting whether an image and an answer match, and 3) predicting answers solely based on questions. 
\item On our PathVQA dataset, we demonstrate the effectiveness of our proposed learning-by-ignoring and cross-modal SSL methods in detecting noisy training examples and learning powerful visual-textual representations.
\end{itemize}

\vspace{-0.3cm}
\section{Related Works}
\vspace{-0.2cm}
\subsection{Medical VQA Datasets}
\vspace{-0.2cm}

To our best knowledge, there are two existing datasets for medical visual question
answering.
 The VQA-Med~\citep{medicalVQA_dataset19} dataset is created on 4,200 radiology images and has 15,292 question-answer pairs. Most of the questions are in multiple-choice (MC) style and can be answered by multi-way classifiers. 
 This makes the difficulty of this dataset significantly lower. 
 VQA-RAD~\citep{nature} is a manually-crafted dataset where questions and answers are given by clinicians on radiology images. It has 3515 questions of 11 types.
Our dataset  differs from VQA-Med and VQA-RAD in two-fold. First, ours is about pathology while VQA-Med and VQA-RAD~\citep{nature} are both about radiology. Second, our dataset is a truly challenging QA dataset where most of the questions are open-ended while in VQA-Med and VQA-RAD the majority of questions have a fixed  number of  candidate answers and can be answered by multi-way classification. Besides, the number of questions in our dataset is much larger than that in VQA-Med and VQA-RAD.

\vspace{-0.3cm}
\subsection{Self-supervised Learning}
\vspace{-0.2cm}

 \begin{wrapfigure}{r}{0.6\textwidth}
\vspace{-0.4cm}
    \centering
        \includegraphics[width=0.6\columnwidth]{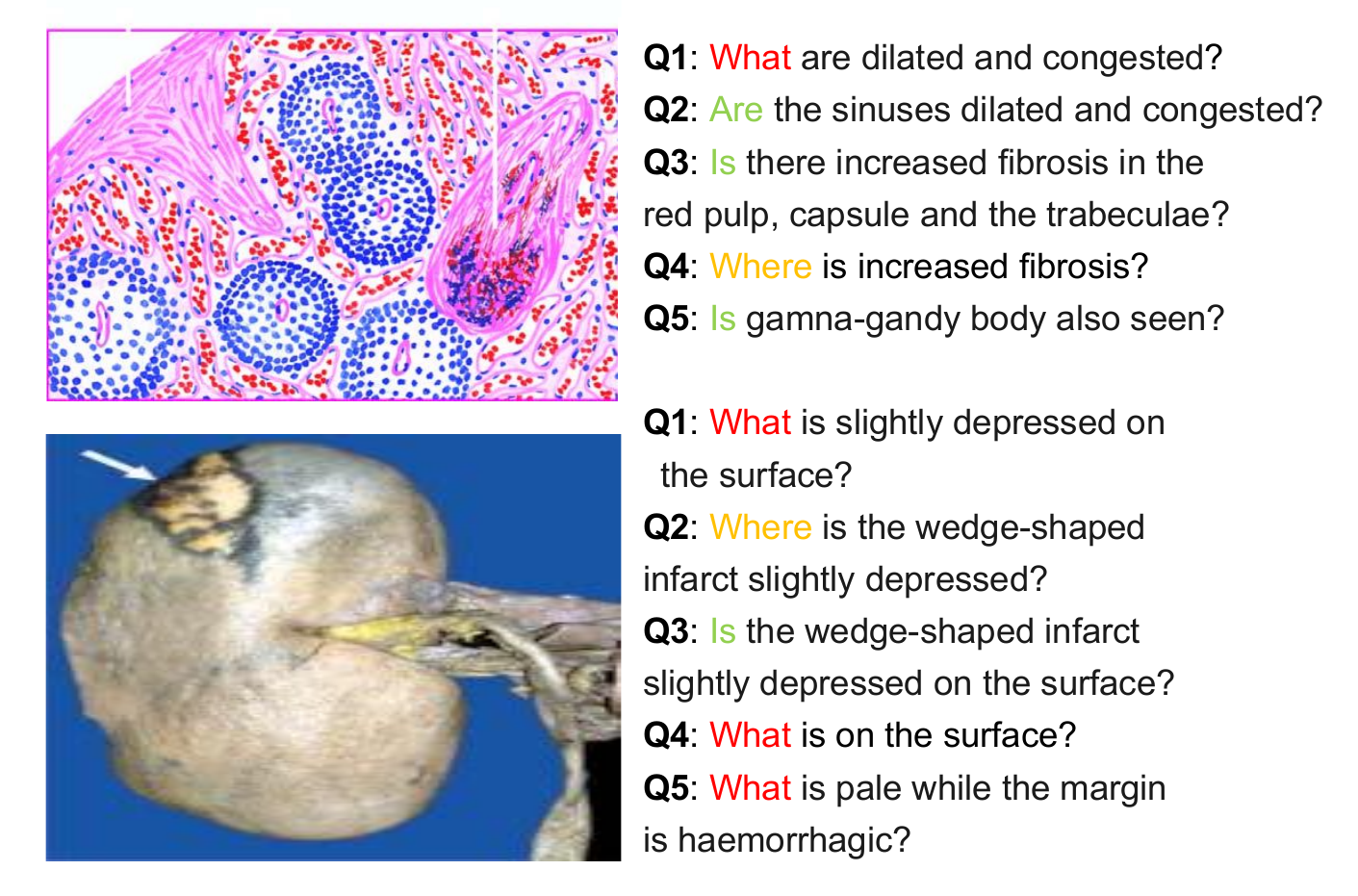}
        \vspace{-0.6cm}
       \caption{Two exemplar images with  generated questions. Both images have three types of questions: ``what", ``where", and ``yes/no".}
       \vspace{-0.4cm}
        \label{illustrate}
\end{wrapfigure}Self-supervised learning (SSL) has been widely studied to learn better representations of images and texts. 
SSL learns useful features automatically by constructing
a loss from a pretext task  without much demand for human annotations. It purely uses the input data to create auxiliary tasks and enables deep networks to learn effective latent features by solving these auxiliary tasks.  
Various strategies have been proposed to construct auxiliary tasks, based on temporal correspondence~\citep{li2019joint,wang2019learning}, cross-modal consistency~\citep{wang2019reinforced}, etc. In computer vision, examples of auxiliary tasks include rotation prediction~\citep{gidaris2018unsupervised}, image inpainting~\citep{pathak2016context}, automatic colorization~\citep{zhang2016colorful}, instance discrimination~\citep{wu2018unsupervised}, to name a few. In SSL for natural language processing, examples of auxiliary tasks include next-word prediction in the GPT model~\citep{gpt2}, next sentence prediction, masked word prediction in the BERT model~\citep{bert}, and so on.

Cross-modal self-supervised learning has been studied as well, which learns representations for data with multiple modalities by solving cross-modal auxiliary tasks.  
VisualBERT~\citep{visualbert} learns representations for images and texts by implicitly aligning elements of an input text and regions in an associated input image with self-attention. Two visually-grounded language model objectives are used for pretraining VisualBERT on image caption data.

VideoBERT~\citep{videobert} performs vector quantization on video frames to get visual tokens, and then trains masked language models on the concatenation of visual tokens and text tokens. \citet{chung2020perfect} proposes to learn cross-modal joint embeddings using self-supervised learning for cross-modal retrieval. CVLP~\citep{Visual_Linguistic} proposes an  unbiased
contrastive visual-linguistic pretraining approach, which constructs a visual self-supervised loss based on contrastive learning. 
LXMERT~\citep{lxmert} designs five pretraining tasks: masked language
modeling, feature regression, label classification,
cross-modal matching, and image question answering, to
pretrain a large Transformer model.
ViLBERT~\citep{lu2019vilbert} proposes to pretrain a vision-and-language BERT model through masked multi-modal modeling and multi-modal alignment prediction tasks and then transfer
the model to visual question answering tasks.

\vspace{-0.3cm}
\subsection{Data selection and data reweighting}
\vspace{-0.2cm}
A number of approaches have been proposed for data selection. Matrix column subset selection~\citep{deshpande2010efficient,boutsidis2009improved} aims to select a subset of data examples that can best reconstruct the entire dataset. Similarly, coreset selection~\citep{bachem2017practical} chooses representative training examples in a way that models trained on the selected examples have comparable performance with those trained on all training examples.  These methods perform data selection and model training separately.  As a result, the validation performance of the model cannot be used to guide data selection. 
\citet{ren2018learning} propose a meta learning method to learn the weights of  training examples by performing a meta gradient descent step on the weights of the current mini-batch of examples. \begin{wraptable}{r}{0.5\textwidth}
    \centering
    \vspace{-0.7cm}
\caption{Statistics of the PathVQA dataset}
 \begin{tabular}{lccc}
\toprule
& Max& Avg& Min  \\
\hline 
\# questions per image& 14&6.6  & 1   \\
\hline
\# words per question& 28& 9.5& 3 \\
\hline
\# words per answer & 10& 2.5   &1  \\
\bottomrule
\end{tabular}
\vspace{-0.3cm}
\label{Statistics result}
\end{wraptable} \citet{shu2019meta} propose a method which can adaptively learn an explicit weighting function
directly from data. 
Different from these works, our learning-by-ignoring method is based on a bi-level optimization framework which can flexibly select data elements with various granularity, such as pixels, images, bags of instances, etc., in a unified way.
\vspace{-0.2cm}

\vspace{-0.1cm}
\section{The PathVQA Dataset}
\vspace{-0.2cm}
The PathVQA dataset consists of 32,795 question-answer pairs generated from 1,670 pathology images collected from two pathology textbooks: ``Textbook of Pathology"~\citep{muir1941text} and ``Basic Pathology"~\citep{robbins1981basic}, and 3,328 pathology images collected from the PEIR\footnote{\url{http://peir.path.uab.edu/library/index.php?/category/2}} digital library.  \begin{wraptable}{r}{0.4\textwidth}
    \centering
    \vspace{-0.6cm}
\caption{Frequency of questions in different categories}
    \scalebox{0.8}{
\begin{tabular}{ll}
    \toprule 
  \multirow{2}*{Question type} & Total number   \\
 & and percentage \\
    \hline
    Yes/No & 16,329 (49.8\%)\\
    \hline
    What&13,401 (40.9\%) \\
    \hline
    Where&2,157 (6.6\%) \\
    \hline                                                        
    How& 595 (1.8\%)\\
    \hline
     How much/many& 139 (0.4\%)  \\
    \hline
    Why& 114 (0.3\%)  \\
    \hline
    When&51 (0.2\%)\\
    \hline
    Whose& 9 (0.1\%)  \\
    \bottomrule
    \end{tabular}
    }
\vspace{-0.3cm}
\label{qresult}
\end{wraptable} Figure~\ref{illustrate} shows some examples. On average, each image has 6.6 questions. The maximum and minimum number of questions for a single image is 14 and 1 respectively. The average number of words per question and per answer is 9.5 and  2.5 respectively. Table~\ref{Statistics result} summarizes these statistics.  There are eight  different categories of questions: what, where, when, whose, how, why, how much/how many, and yes/no. Table~\ref{qresult} shows the number of questions and percentage of each category. The questions in the first 7 categories are open-ended: 16,466 in total and accounting for 50.2\% of all questions. The rest are close-ended ``yes/no" questions. The questions cover various aspects of  visual contents, including color, location, appearance, shape, etc. Such clinical diversity poses great  challenges for AI models to solve this pathology VQA problem.

\vspace{-0.4cm}
\section{Methods}
\vspace{-0.3cm}
In this section, we propose a learning-by-ignoring approach for automatically identifying and removing problematic training examples to avoid distorting the model by these bad-quality examples. We also propose several cross-modal self-supervised learning methods to learn effective visual and textual representations. 
These proposed methods can be applied to any VQA method. In this work, we choose two well-established and state-of-the-art VQA methods to perform the study while noting that other VQA methods are applicable as well.

\vspace{-0.4cm}
\subsection{Learning to Ignore}
\vspace{-0.3cm}
To automatically identify and remove bad-quality examples from the training data to avoid distorting the model by them, we propose a learning-by-ignoring (LBI) approach, where 
a data example is taken as the input and a corresponding ignoring variable $a\in[0,1]$ is learned to indicate how likely this example should be ignored.
For the loss $L$ defined on each training example, we multiply it with the ignoring variable. If $a$ is close to zero, then $L$ is close to zero and this data example does not contribute to model training. We learn these ignoring variables using the following formulation:
\begin{equation}
\begin{array}{ll}
\textrm{min}_A & \sum_{i=1}^{N^{(\mathrm{val})}}  L(d_{i}^{(\mathrm{val})}; W^{*}(A))
\\ \text {s.t.} & W^{*}(A)=\operatorname{argmin}_{W} \;\; \sum_{i=1}^{N^{(\mathrm{tr})}} a_i L(d_{i}^{(\mathrm{tr})}; W)\end{array}
\end{equation}
where $A=\{a_i\}_{i=1}^{N^{(\mathrm{tr})}}$. $W$ denotes the weights of the VQA model. $L(d_{i}^{(\mathrm{tr})}; W)$ is the training loss defined on the training example $d_{i}^{(\mathrm{tr})}$. $a_{i} \in [0,1]$ is an ignoring variable indicating how likely $d_{i}^{(\mathrm{tr})}$ should be ignored. 
Given the weighted training loss $\sum_{i=1}^{N^{(\mathrm{tr})}} a_i L(d_{i}^{(\mathrm{tr})}; W)$, we learn the VQA model weights $W$ by minimizing this loss and get the optimal weights $W^*$. Note that $W^*$ is a function of $A$. When $A$ changes, the ignoring variables changes and the weighted training loss changes. The optimal model trained by minimizing the weighted training loss changes accordingly. Given the trained VQA model $W^{*}(A)$, we measure its loss on the validation dataset $\sum_{i=1}^{N^{(\mathrm{val})}}  L(d_{i}^{(\mathrm{val})}; W^{*}(A))$. We assume all validation examples are double-checked by humans and have good quality. \begin{wrapfigure}{r}{0.5\textwidth}
\vspace{-0.3cm}
    \centering
 \includegraphics[width=0.5\textwidth]{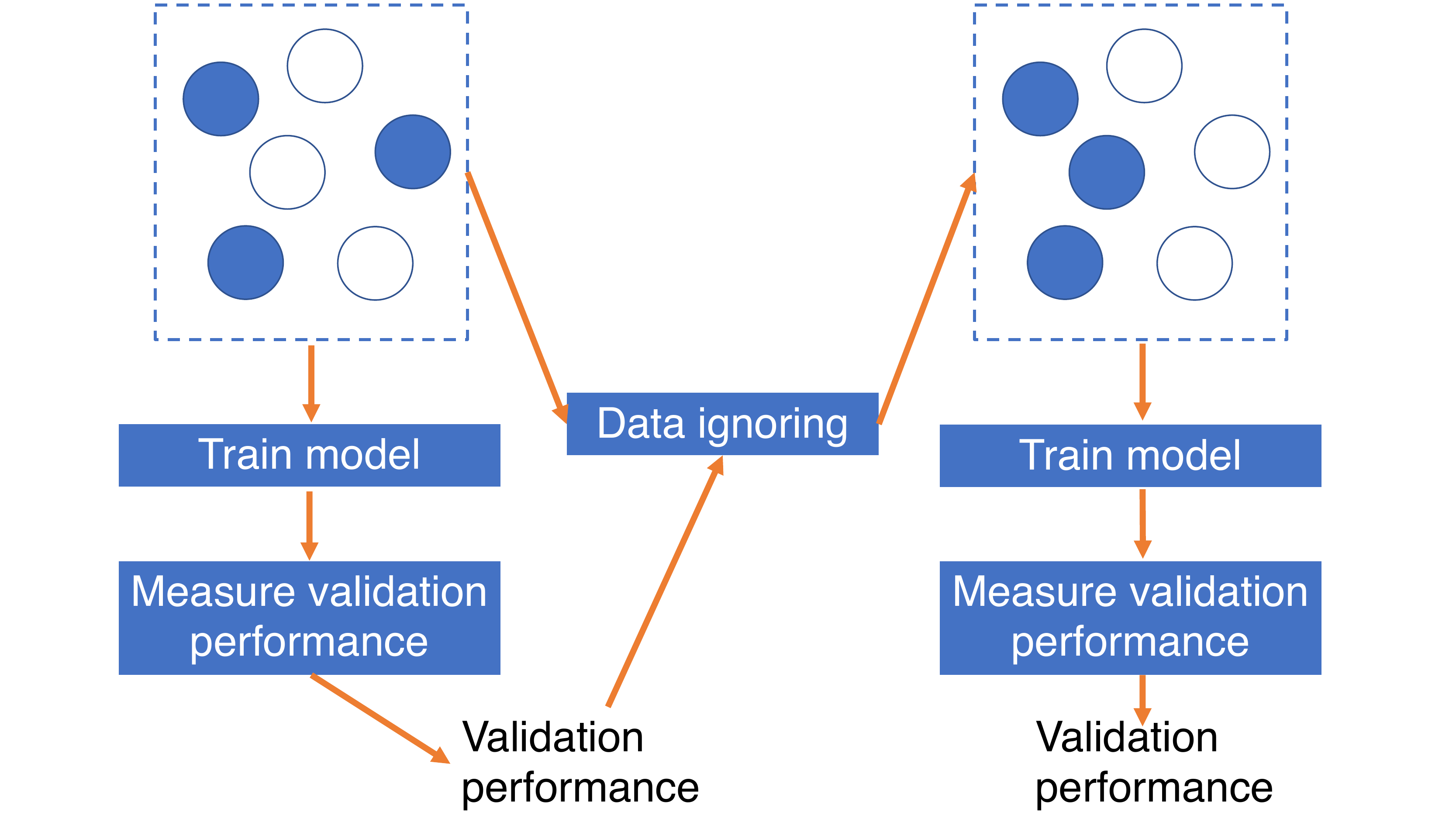}
 \vspace{-0.5cm}
       \caption{Learning by ignoring.}
       \vspace{-0.5cm}
 \label{illustrate learning by ignoring}
\end{wrapfigure} The validation loss is a function of $A$. We learn $A$ by minimizing this validation loss, i.e., finding the optimal ignoring variables to remove bad-quality training examples so that the model trained on the remaining good-quality examples achieves the best performance on the validation set. Figure~\ref{illustrate learning by ignoring} illustrates the idea. In our PathVQA dataset, the number of training data examples is not very large (tens of thousands), we can directly learn an ignoring variable for each data example. In other applications, if there are millions of training examples, learning millions of ignoring variables may not be a good choice. Under such circumstances, we can use a neural network (called ignoring network) to parameterize the ignoring variable, where the input of the network is a feature representation of the data example and the output of the network is an ignoring variable. The ignoring network and the VQA model can share the same encoder used for representation learning. 
\begin{algorithm}[h]
\caption{Algorithm for learning-by-ignoring}      
\label{alg1}                       
\begin{algorithmic}
 \STATE
 \WHILE{not converged}
\STATE 1. Update ignoring variables $A$ by descending  $\nabla_{A} L_{v a l}\left(W-\xi \nabla_{W} L_{train }(W, A) \right)$
\STATE 2. Update weights $W$ by descending $\nabla_{W} L_{train }(W, A)$
\ENDWHILE
\end{algorithmic}
\end{algorithm}
\vspace{-0.3cm}

The algorithm of learning-by-ignoring  is shown in Algorithm~\ref{alg1}. Similar to~\citet{liu2018darts}, we approximate $W^*(A)$ using one step of gradient descent update of $W$: $W^*(A)=W-\xi \nabla_{W} L_{train }(W, A)$ where $L_{train }(W, A)=\sum_{i=1}^{N^{(\mathrm{tr})}} a_i L(d_{i}^{(\mathrm{tr})}; W)$. Then we plug this approximation into the validation loss: $L_{v a l}\left(W-\xi \nabla_{W} L_{train }(W, A) \right)$, and update $A$ by performing gradient descent on the approximated validation loss. The update of $W$ and $A$ are performed alternatively until convergence.

\vspace{-0.3cm}
\subsection{Self-supervised learning on PathVQA}
\vspace{-0.2cm}
\begin{figure}[ht]
\vspace{-0.2cm}
    \centering
 \includegraphics[width=0.9\textwidth]{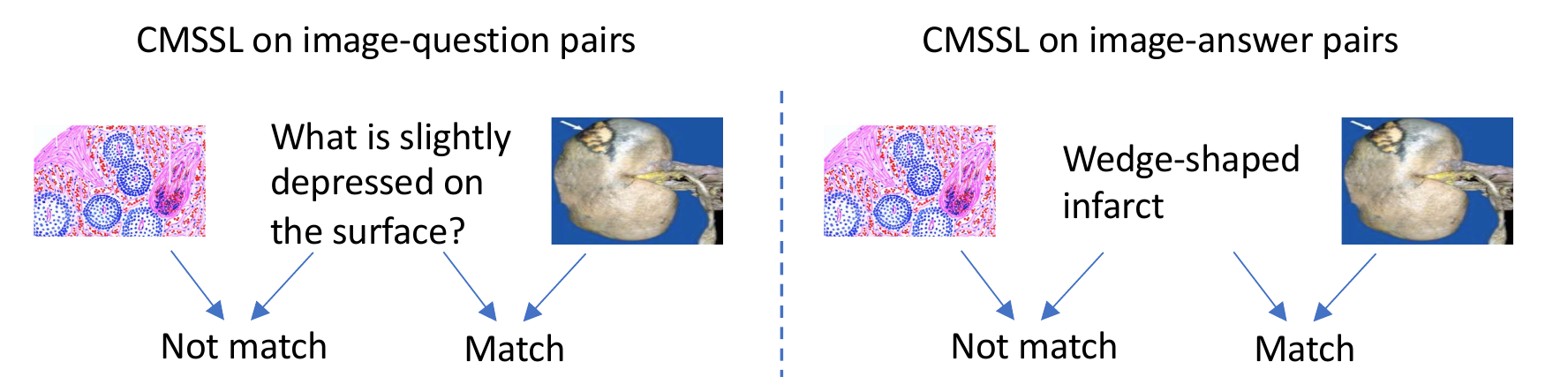}
 \vspace{-0.4cm}
       \caption{Cross-modal self-supervised learning.}
      \vspace{-0.3cm}
 \label{illustrate image-q}
\end{figure}
To learn powerful visual and textual representations on limited data, 
we develop cross-modal self-supervised learning (CMSSL) approaches.
Given a pair of data $(x,y)$, where $x$ and $y$ are from different modalities and could be an image, a question, or an answer, we define a CMSSL task on $(x,y)$ which judges whether $x$ and $y$ are from the same training example. We learn the image encoder and text encoder by solving this task. 
Figure~\ref{illustrate image-q} illustrates the idea. 
We define this task on two types of pairs: image-question and image-answer. For image-question CMSSL, given an image and a question, if the question is asked based on the content of the image, then they are labeled as a match. Otherwise, they are labeled as not a match. The image encoder and question encoder are learned jointly to predict whether an image and a question match with each other. Such a task can help to learn the correspondence between image regions and words in the question. 
The encoders learned in CMSSL are used to initialize the encoders in VQA models, which are then finetuned on the PathVQA dataset. 
We also perform CMSSL on image-answer pairs: given an image and an answer, judge whether this answer is relevant to the image. We use the question encoder to encode the answer. 
In addition, we perform single modal SSL on question-answer pairs (SSL-QA), 
to capture the semantic correspondence between words in questions and answers. In SSL-QA, we first pretrain the question encoder by predicting answers solely based on questions themselves without using image information, then finetune the question encoder together with other modules in the full VQA task involving images. Given CMSSL on image-question pairs, CMSSL on image-answer pairs, and SSL-QA, we perform them simultaneously in a multi-task learning framework which minimizes the weighted sum of losses of the three SSL tasks.

\vspace{-0.3cm}
\subsection{VQA Models}
\vspace{-0.2cm}
We evaluate the effectiveness of learning-by-ignoring and cross-modal self-supervised learning on two VQA models. 
\begin{itemize}[leftmargin=*]
\vspace{-0.3cm}
\item \textbf{Method 1}:
In~\citet{lxmert}, a large-scale Transformer~\citep{vaswani2017attention} 
model is built that consists of three encoders: an object
relationship encoder, a language encoder,
and a cross-modal encoder. The three encoders are built mostly based on two kinds of attention layers --- self-attention layers and cross-attention layers. The object
relationship encoder and the language encoder are both single-modality encoders. Each
layer of them contains a self-attention sub-layer and a feed-forward sub-layer, where the feed-forward sub-layer is composed
of two fully-connected sub-layers. A  cross-modal
encoder is proposed to learn the connections between
vision and language. Each layer of it consists of two self-attention
sub-layers, one bi-directional cross-attention sub-layer,
and two feed-forward sub-layers.

\item \textbf{Method 2}:    
The method proposed in~\citet{BAN} uses a Gated Recurrent Unit (GRU)~\citep{GRU} recurrent network and a Faster R-CNN~\citep{RCNN} network to embed the question and the image.
 It extends the idea of co-attention into bilinear attention which considers every pair
of multimodal channels. It learns bilinear attention distributions using the bilinear attention networks (BAN) and uses low rank approximation techniques to approximate the bilinear interaction between question embeddings and  image embeddings. It also proposes residual learning of attention which keeps
the size of intermediate features constant.
 \vspace{-0.3cm}
\end{itemize}

\begin{wraptable}{r}{0.5\textwidth}
    \centering
    \vspace{-0.7cm}
\caption{Statistics of the data split}
    \scalebox{0.9}{
 \begin{tabular}{lccc}
\toprule
& Training set& Validation set& Test set \\
\hline 
\# images& 3,021& 987  & 990  \\
\# QA pairs&19,755 &6,279 &6,761 \\
\bottomrule
\end{tabular}
}
\vspace{-0.5cm}
\label{data_in_each_set}
\end{wraptable}

\vspace{-0.1cm}
\section{Experiment}
\vspace{-0.4cm}
\subsection{Experimental Settings}
\vspace{-0.3cm}
\paragraph*{Data split}
We partition the images in the PathVQA dataset along with the associated questions into a training set, validation set, and testing set with a ratio of about 3:1:1. In the PathVQA dataset, the frequencies of question categories are imbalanced. Because of this, during the partition process, we perform sampling to ensure the frequencies of these categories in each set to be consistent. There are 19,755 question-answer pairs in the training set, 6,279 in the validation set, and 6,761 in the testing set. For all the data examples in the validation set and test set, senior radiologists helped to carefully examine them to ensure they are clinically correct. The training set was not examined by senior radiologists. 
The statistics are summarized in
 Table~\ref{data_in_each_set}.

\vspace{-0.4cm}
\paragraph*{Implementation details}
We basically follow the original model configurations used in~\citet{lxmert}, ~\citet{BAN}, and~\citet{SAN}. Data augmentation is applied to the images, including shifting, scaling, and shearing.  From questions and answers in the PathVQA dataset, we create a vocabulary of 4,631 words that have the highest frequencies.
\begin{table}[t]
    \centering
\caption{Accuracy (\%), BLEU-$n$ (\%), and F1 (\%) achieved by different methods. We denote cross-modal SSL on image-question pairs and image-answer pairs as CMSSL-IQ, CMSSL-IA, and denote single-modal SSL on question-answer pairs as SSL-QA}
\begin{tabular}{lcccccc}
\toprule 
Method & Accuracy  &BLEU-1&BLEU-2 &BLEU-3 & F1\\
\hline
Method 1 without image
& 49.2 &50.2& 2.8&1.2 &9.5 \\
Method 1
& 57.6& 57.4&3.1& 1.3&9.9 \\
Method 1 with ignoring
& 58.5 &58.9 &3.5&2.0 &10.2 \\
Method 1 with CMSSL-IQ
& 58.7& 59.0& 3.5& 2.1&11.0\\
Method 1 with CMSSL-IA
& 58.6&58.9& 3.4&2.0 &10.3\\
Method 1 with SSL-QA
& 58.7 & 59.0&3.5&2.1&11.2\\
Method 1 with joint pretraining
& 59.3&59.2&4.7&2.8 &11.6\\
\rowcolor{mygray} Method 1 with joint pretraining+ignoring
&  60.1&59.9&5.1&3.2&12.2\\
\hline 
Method 2 without image
& 46.2&46.5 &1.0 &0.0 & 0.8\\
Method 2
& 55.1&56.2&3.2&1.2&8.4 \\
Method 2 with ignoring
& 56.3&57.4 & 3.5&1.8 &9.6\\
Method 2 with CMSSL-IQ 
& 55.9&57.1&3.4&1.4& 9.2\\
Method 2 with CMSSL-IA
& 55.9&57.1&3.5&1.5&9.2\\
Method 2 with SSL-QA     
& 57.6&58.8&4.1&1.5&10.8\\
 Method 2 with joint pretraining
& 57.7 &59.1&4.2&2.2&10.9 \\
\rowcolor{mygray} Method 2 with joint pretraining+ignoring
&  58.4 &59.5&4.4&2.6&11.2\\
\bottomrule
\end{tabular}
\label{tab:test result}
\vspace{-0.5cm}
\end{table}
In Method 1, we use the default hyperparameter settings in~\citet{lxmert}. For the text encoder, the hidden size was set to 768. The image features were extracted from the outputs of the Faster-RCNN network, which is pretrained on BCCD\footnote{https://public.roboflow.ai/object-detection/bccd} -- a medical dataset containing blood cells photos, as well as on Visual Genome~\citep{visualg}. The initial learning rate was set to 5e-5 with the Adam~\citep{adam} optimizer used. The batch size was set to 256. The model was trained  for 200 epochs.
In the cross-modal SSL pretraining on Method 1, we train a  linear classifier with a dimension of 1,280 to judge whether one modality of data (image, question, answer) matches with another. 
In Method 2, words in questions and answers are represented using GloVe~\citep{glove} vectors pretrained on general-domain corpora such as Wikipedia, Twitter, etc. The image features are extracted from the outputs of the Faster-RCNN network pretrained on BCCD and Visual Genome. Given an image and a question, the model outputs an answer from a predefined set of answers. The dropout~\citep{dropout} rate for the linear mapping was set to 0.2 while for the classifier it was set to 0.5. The initial learning rate was set to 0.005 with the Adamax optimizer~\citep{kingma2014adam} used. The batch size was set to 512. The model was trained for 200 epochs.
In the cross-modal SSL pretraining on Method 2, similar to that on Method 1, we train a linear classifier with a dimension of 1,280 to predict whether two modalities of data match or not. For learning-by-ignoring, we update ignoring variables using the Adam optimizer, with an initial learning rate of 0.01. We perform training for 120 epochs in Method 1 with ignoring and for 180 epochs in Method 2 with ignoring. 

\vspace{-0.4cm}
\paragraph*{Evaluation metrics}
We perform evaluation using three metrics: 1) accuracy~\citep{first} which measures  the  percentage of inferred answers that match exactly with the ground-truth using string matching; only exact matches are considered as correct; 2) macro-averaged F1~\citep{F1}, which measures the average overlap between the predicted answers and ground-truth, where the answers are treated as bag of tokens; 3) BLEU~\citep{bleu}, which measures the similarity of  predicted answers and ground-truth by matching $n$-grams.

\vspace{-0.3cm}
\subsection{Results}
\vspace{-0.2cm}

Table~\ref{tab:test result} shows the VQA performance achieved by different methods. From this table, we make the following observations. \textbf{First}, for both Method 1 and Method 2, applying learning-by-ignoring (LBI) improves the performance. This demonstrates the effectiveness of LBI in improving the generalization ability of trained VQA models. LBI learns to identify and remove noisy and erroneous training data examples, which can avoid the model to be distorted by such bad-quality examples. 
\textbf{Second}, for both Method 1 and 2, applying cross-modal SSL (CMSSL) methods including CMSSL-IQ and  CMSSL-IA improves the performance, which demonstrates the effectiveness of CMSSL.  CMSSL uses auxiliary tasks, including judging whether an image matches with a question and judging whether an image matches with an answer, to learn semantic correspondence between image regions and words in questions/answers, which can improve the effectiveness of visual and textual representations for accurate VQA. \textbf{Third}, using SSL-QA improves VQA performance of Method 1 and 2. SSL-QA learns the correspondence between words in questions and words in answers, which can better extract semantic representations of questions and answers. 
\textbf{Fourth}, joint pretraining which performs CMSSL-IQ, CMSSL-IA, and SSL-QA jointly achieves better performance than performing the three SSL tasks individually, for both Method 1 and 2. This is because letting the model solve several SSL tasks simultaneously is more challenging, which encourages  the model to learn more powerful textual and visual representations. \textbf{Fifth}, applying both joint pretraining and learning-by-ignoring achieves the best performance in  Method 1 and 2. 
\begin{wraptable}{r}{0.47\textwidth}
\vspace{-0.6cm}
    \centering
\caption{Accuracy (\%) on ``yes/no" questions}
    \scalebox{0.8}{
  \begin{tabular}{lc}
    \toprule 
 Method&Accuracy \\
    \hline 
Method 1 without image&85.1\\
Method 1 & 86.1\\
Method 1 with ignoring
& 86.4\\
Method 1 with CMSSL-IQ
& 86.2\\
 Method 1 with CMSSL-IA
& 86.4\\
Method 1 with SSL-QA
& 86.2\\
Method 1 with joint pretraining
&  86.8\\
\rowcolor{mygray} Method 1 with joint pretraining+ignoring
& 87.1 \\
\hline
Method 2 without image  &84.5 \\
Method 2 
&  85.7\\
Method 2 with ignoring
& 86.4\\
Method 2 with CMSSL-IQ 
& 86.4\\
Method 2 with CMSSL-IA
& 86.4\\
 Method 2 with SSL-QA
& 86.8\\
Method 2 with joint pretraining
& 86.6 \\
\rowcolor{mygray} Method 2 with joint pretraining+ignoring
& 87.2 \\
   \bottomrule
    \end{tabular}
    }
\vspace{-0.3cm}
\label{yes/no-acc}
\end{wraptable} Table~\ref{tab:question types result} shows the accuracy scores achieved on open-ended questions belonging to the following categories: what, where, how, how much/how many, and why respectively by different methods. Table~\ref{yes/no-acc} shows the accuracy on yes/no questions.
 Similar to the observations made from Table~\ref{tab:test result}, the results in  Table~\ref{tab:question types result} and Table~\ref{yes/no-acc} also demonstrate that  learning-by-ignoring and cross-modal SSL both help to improve VQA performance.  In Table~\ref{tab:question types result},   all  methods perform the best on ``where" questions. This is because it is relatively easy to recognize image regions of interest for ``where" questions, which helps the model to give the correct answer. In Table~\ref{yes/no-acc}, all methods perform much better than random guesses (where the accuracy is 50\%). This indicates that our PathVQA dataset is clinically meaningful, which allows VQA models to be learnable.

\begin{table*}[t]
\small
    \centering
    \vspace{-0.8cm}
\caption{Accuracy (\%) on open-ended questions of different types}
\begin{tabular}{lccccc}
\toprule
\multirow{2}{*}{Method}&\multicolumn{5}{c}{Question types }    \\
\cline{2-6}
& What& Where& How &How much/many  &Why \\
\hline 
Method 1 without image&0.08&0.39&0.16&0.41&0.50\\
Method 1 &0.22 & 0.73  &0.12&0.45 & 0.50\\
Method 1 with ignoring &0.24&0.76&0.15&0.45&0.64\\
Method 1 with CMSSL-IQ& 0.24 & 0.73&0.13&0.45&0.59
\\
Method 1 with CMSSL-IA &0.24&0.74&0.13&0.45&0.59
\\
Method 1 with SSL-QA &0.26&0.78&0.15&0.50&0.64\\
Method 1 with joint pretraining
&0.29&0.79&0.16&0.50&0.68  \\
Method 1 with joint pretraining+ignoring
&  \textbf{0.32}&\textbf{0.81}&\textbf{0.16}&\textbf{0.56}&\textbf{0.68}\\
\hline
Method 2 without image & 0.05&0.29&0.00&0.00&0.00\\
Method 2 &0.18 &  0.64& 0.11 & 0.36 &0.32 \\
Method 2 with ignoring & 0.24 &0.72&0.12&0.41&0.41\\
Method 2 with CMSSL-IQ &0.20&0.71&0.12&0.36&0.50\\
Method 2 with CMSSL-IA&0.20 &0.72&0.11&0.41&0.45\\
Method 2 with SSL-QA &0.20&0.71&0.11&0.36&0.45\\
Method 2 with joint pretraining
& 0.21&0.72&0.12&0.45&0.55 \\
Method 2 with joint pretraining+ignoring
& \textbf{0.24}&\textbf{0.72}&\textbf{0.14}&\textbf{0.45}&\textbf{0.59} \\
\bottomrule
\end{tabular}
\label{tab:question types result}
\vspace{-0.5cm}
\end{table*}

One may suspect how much information in images are used during the inference of the answers? Could it be possible that the models simply learn the correlations between questions and answers and ignore the images? In light of these concerns, we perform studies where the images are not fed into VQA models and only questions are used as inputs for inferring answers. Table~\ref{tab:test result} shows the results of not using images (``Method 1/2 without image”). As can be seen, for both Method 1 and 2, ignoring images leads to substantial degradation of performance. This shows that images in our dataset provide valuable information for VQA and PathVQA is a meaningful VQA dataset. The models trained on our datasets are not degenerated to simply capture the correlation between questions and answers. %

\vspace{-0.3cm}
\section{Conclusion}
\vspace{-0.3cm}
In this paper, towards the goal  of developing AI systems to pass the board-certificated examinations of the American Board  of Pathology and fostering research in medical visual question answering, we build a pathology VQA dataset -- PathVQA -- that contains 32,795 question-answer pairs of 8 categories, generated from 4,998 images. Majority of questions in  our dataset are open-ended, posing great challenges for the medical VQA research. Our dataset is publicly available. To address the challenges that the training data may contain errors and the effective representations of pathology images and questions are difficult to learn on limited data, we propose a learning-by-ignoring approach to automatically identify and remove problematic training examples and develop  cross-modal self-supervised learning approaches to learn visual and textual representations effectively. 
The experiments on our collected PathVQA dataset demonstrate the effectiveness of our proposed methods.

\bibliography{iclr2021_conference}
\bibliographystyle{iclr2021_conference}

\clearpage
\appendix
\section{Appendix}

\subsection{Example of ABP test questions} 
\begin{figure}[H]
	\begin{center}
 	\includegraphics[width = 0.7\columnwidth]{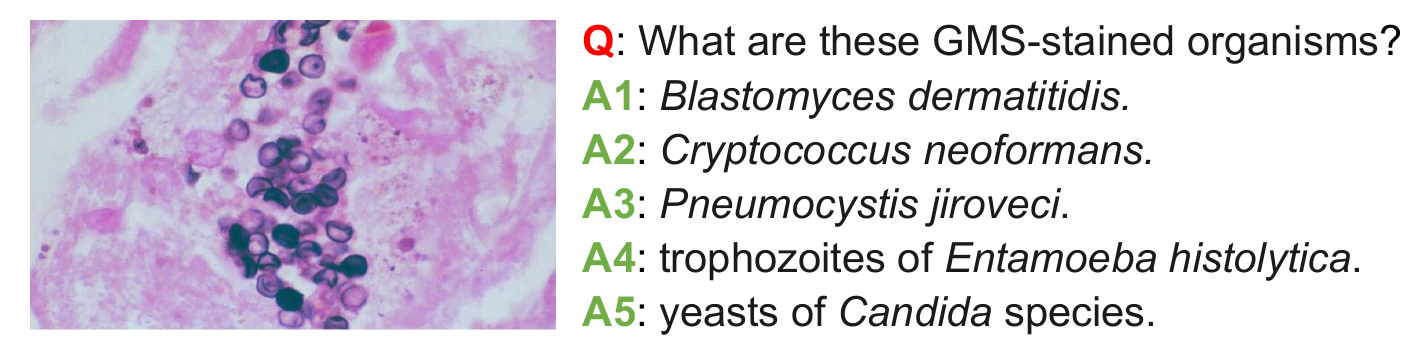}
 	\caption{ An example of ABP test questions. 
	}\label{ABP}
	\end{center}
 \end{figure}

An example of ABP test questions is shown in Figure~\ref{ABP}.  

\subsection{Comparison of existing VQA datasets} 
 The comparison of existing VQA datasets is shown in Table~\ref{dataset}. Table~\ref{dataset} presents a comparison of different VQA datasets. The first five datasets are in the general domain while the last three are in the medical domain. Not surprisingly, the size of general-domain datasets (including the number of images and question-answer pairs) is much larger than that of medical datasets since general-domain images are much more available publicly and  there are many qualified human annotators to generate QA pairs on general images. Our dataset is larger than the two medical datasets: VQA-Med and VQA-RAD, and majority of questions in our dataset are open-ended while majority of questions in VQA-Med and VQA-RAD are in multiple-choices style.
 \begin{table}[H]
 \centering
\caption{Comparison of VQA datasets}
\begin{tabular}{|c|c|c|c|c|}
\hline  
&Domain&\# images& \# QA pairs &Answer type\\
\hline  
DAQUAR&General& 1,449&12,468  & Open \\
\hline
VQA&General& 204K& 614K & Open/MC \\
\hline
VQA v2 &General& 204K& 1.1M & Open/MC \\ 
\hline
COCO-QA & General&123K& 118K & Open/MC \\
\hline
CLEVR&General& 100K& 999K & Open \\ 
\hline
\hline
VQA-Med &Medical& 4,200& 15,292 & Open/MC \\ 
\hline
VQA-RAD &Medical& 315& 3,515 & Open/MC \\ 
\hline
Ours&Medical& 4,998& 32,795 & Open \\ 
\hline
\end{tabular}
\label{dataset}
\end{table}

\subsection{Number of questions in different categories for training, validation, and test set}
For our data split, the number of questions in different categories in each set  is shown in Table~\ref{tab:question type number}. 
\begin{table}[H]
    \centering
\caption{Number of questions in different categories in each set}
\begin{tabular}{ccccccc}
\toprule  
\multirow{2}{*}{Dataset}&\multicolumn{6}{c}{Question types }    \\
\cline{2-7}
& What& Where& How &How much/many  &Why &Yes/No\\
\hline
Training set&8083 & 1316 & 366 &62& 71&9804\\
Validation set&2565 &409  &108  & 21 &21& 3135\\
Testing set &2753 & 432 &121&18 & 22&3390\\
\bottomrule
\end{tabular}
\label{tab:question type number}
\end{table}

\subsection{Derivation of Gradient in algorithm~\ref{alg1}} \label{sec:grad}
In Algorithm~\ref{alg1}, there are two steps. In the first step, we
 update ignoring variables $A$ by descending \\ $\nabla_{A} L_{v a l}\left(W-\xi \nabla_{W} L_{train }(W, A) \right)$, where we approximate $W^*$ using one step gradient descent update of $W$: $$W^* \approx W-\xi \nabla_{W} L_{train }(W, A),$$
 where $\xi$ is the learning rate.

We compute $\nabla_{A} L_{v a l}\left(W^{*}\right)$ as follows:
\begin{subequations}
\begin{align}
& \quad \nabla_{A} L_{v a l}\left(W^{*}\right )
  \\&\approx \nabla_{A} L_{v a l}\left(W-\xi \nabla_{W} L_{train }(W, A)\right) 
  \\&= -\xi \nabla_{A, W}^{2} L_{train}\left(W, A\right) \nabla_{W^{*}} L_{v a l}\left(W^{*}\right) 
  \\& \approx-\xi\frac{\nabla_{A} L_{train}\left(W^{+}, A\right)-\nabla_{A} L_{train}\left(W^{-}, A\right)}{2 \epsilon},
  \end{align}
\end{subequations}
where $\epsilon$ is a small scalar and
$$W^{\pm}=W \pm \epsilon \nabla_{W^{*}} L_{val}\left(W^{*}\right).$$

\subsection{Additional Related Works}
A number of visual question answering datasets have been developed in the general domain. DAQUAR~\citep{first} is built on top of the NYU-Depth V2 dataset~\citep{nyu} which contains RGBD images of indoor scenes. DAQUAR consists of (1) synthetic question-answer pairs that are automatically generated based on textual templates and (2) human-created question-answer pairs produced by five annotators. The VQA dataset~\citep{vqa} is developed on real images in MS COCO~\citep{coco} and abstract scene images
in~\citep{abstract0,abstract1}. The
question-answer pairs are created by human annotators who are encouraged to ask ``interesting" and ``diverse" questions. VQA v2~\citep{goyal} is extended from the VQA~\citep{vqa} dataset to achieve more balance between visual and textual information, by collecting complementary images in a way that each question is associated with a pair of similar images with different answers. In the COCO-QA~\citep{cocoQA} dataset, the  question-answer pairs are automatically generated from image captions based on syntactic parsing and linguistic rules. CLEVR~\citep{clevr,man} is a dataset developed on rendered images of spatially related objects (including cube, sphere, and cylinder)  with different sizes, materials, and colors. The locations and attributes of objects are annotated for each image. The questions are  automatically generated from the annotations.
\end{document}

\maketitle

\section{Code}
The code is availiable on Anonymous Github.
https://anonymous.4open.science/r/f4d98701-c90c-42ab-a98c-80910a0cf05a/

The paths for different methods are listed in Table~\ref{code all}.
\begin{table}[H]
    \centering
    \caption{Code for different methods}
         \centering
    \begin{tabular}{ll}
    \toprule
         Name & Links  \\
          \midrule
Method 1 without image& method1/src/tasks/pvqa.py$^1$\\
Method 1 & method1/src/tasks/pvqa.py$^1$\\
Method 1 with ignoring & method1/src/tasks/pvqa\_ignoring.py$^1$ \\
Method 1 with CMSSL-IQ
& method1/src/pretrain/lxmert\_pretrain.py$^1$\\
Method 1 with CMSSL-IA& method1/src/pretrain/lxmert\_pretrain.py$^1$\\
Method 1 with SSL-QA& method1/src/pretrain/lxmert\_pretrain.py$^1$\\
Method 1 with joint pretraining
&  method1/src/pretrain/lxmert\_pretrain.py$^1$\\
Method 1 with joint pretraining+ignoring
&  method1/src/tasks/pvqa\_ignoring.py$^1$ \\
\hline
Method 2 without image  &method2/finetune\_main.py$^1$ \\
Method 2 
&  method2/finetune\_main.py$^1$\\
Method 2 with ignoring
& method2/finetune\_main\_ignore.py$^1$\\
Method 2 with with CMSSL-IQ 
& method2/pretrain\_main.py$^1$\\
Method 2 with CMSSL-IA
& method2/pretrain\_main.py$^1$\\
Method 2 with SSL-QA
& method2/pretrain\_main.py$^1$\\
Method 2 with joint pretraining
& method2/pretrain\_main.py$^1$ \\
Method 2 with joint pretraining+ignoring
&  method2/finetune\_main\_ignore.py$^1$\\
   \bottomrule
\vspace{-0.3cm}
\label{code all}
    \end{tabular}
        \footnotesize{$^{1}$\url{https://anonymous.4open.science/r/f4d98701-c90c-42ab-a98c-80910a0cf05a/}}}
\end{table}

\section{Experimental setup} 
The data split is performed to ensure the frequencies of question types in training, validation, and testing set to be consistent, and most words of the  vocabulary exist in each set. The ratio of 
training set size, validation set size, and test set size is 3:1:1. 
We implement the methods using PyTorch and perform training on four GTX 1080Ti GPUs.

\section{Hyperparameter settings}
The hyperparameter settings in each experiment are specified in Table~\ref{tab:Baseline hyper-paramter1}-\ref{tab:Baseline hyper-paramter5}. 
\begin{table}[H]
    \centering
    \caption{Hyperparameter settings for Method 1}
    \label{tab:Baseline hyper-paramter1}
    \begin{tabular}{lcl}
    \toprule
         Name & Value & Description \\
         \midrule
         Optimizer & Adam &-\\
         Learning rate &0.00005& Cosine scheduling\\
         \multirow{2}*{Beta }&\multirow{2}*{(0.9, 0.999)}&{Adam coefficients used for computing
            }\\
       \multirow{-2}*{Beta} &\multirow{-2}*{(0.9, 0.999)}&{gradients and their squares}\\
        Max epoch &200&-\\
        Batch size &32&-\\
            \bottomrule
    \end{tabular}
\end{table}

\begin{table}[H]
    \centering
    \caption{Hyperparameter settings for Method 1 with ignoring}
    \label{tab:Baseline hyper-paramter2}
    \begin{tabular}{lcl}
    \toprule
         Name & Value & Description \\
         \midrule
         Optimizer & Adam &-\\
         Optimizer for ignoring variables& Adam &-\\
         Learning rate &0.00005& Cosine scheduling\\
         Learning rate for ignoring variables&0.1&  Weight decay with a rate of 3e-4\\
         \multirow{2}*{Beta }&\multirow{2}*{(0.9, 0.999)}&{Adam coefficients used for computing
            }\\
       \multirow{-2}*{Beta} &\multirow{-2}*{(0.9, 0.999)}&{gradients and their squares}\\
          \multirow{2}*{Beta for ignoring variables }&\multirow{2}*{(0.5, 0.999)}&{Adam coefficients used for computing
            }\\
       \multirow{-2}*{Beta for ignoring variables} &\multirow{-2}*{(0.5, 0.999)}&{gradients and their squares}\\
        Max epoch &120&-\\
        Batch size &16&-\\
            \bottomrule
    \end{tabular}
\end{table}

\begin{table}[H]
    \centering
    \caption{Hyperparameter settings for Method 1 with CMSSL-IQ}
    \label{tab:Baseline hyper-paramter3}
    \begin{tabular}{llcl}
    \toprule
         &Name & Value & Description \\
         \midrule
          \multirow{5} *{Pretraining }   &Optimizer & Adam &-\\
        & Learning rate &0.0001& Cosine scheduling\\
         &\multirow{2}*{Beta}&\multirow{2}*{(0.5, 0.999)}&{Adam coefficients used for computing
            }\\
      & \multirow{-2}*{Beta} &\multirow{-2}*{(0.5, 0.999)}&{gradients and their squares}\\
      &  Max epoch &30&-\\
      &  Batch size &256&-\\
         \midrule
    \multirow{5} *{Finetuning }  &   Optimizer & Adam &-\\
      &   Learning rate &0.00005& Cosine scheduling\\
       &  \multirow{2}*{Beta }&\multirow{2}*{(0.9, 0.999)}&{Adam coefficients used for computing
            }\\
       &\multirow{-2}*{Beta} &\multirow{-2}*{(0.9, 0.999)}&{gradients and their squares}\\
    &    Max epoch &200&-\\
     &   Batch size &32&-\\
            \bottomrule
    \end{tabular}
\end{table}

\begin{table}[H]
    \centering
    \caption{Hyperparameter settings for Method 1 with CMSSL-IA}
    \label{tab:Baseline hyper-paramter4}
    \begin{tabular}{llcl}
    \toprule
        & Name & Value & Description \\
         \midrule
          \multirow{5} *{Pretraining }   &Optimizer & Adam &-\\
        & Learning rate &0.0001& Cosine scheduling\\
         &\multirow{2}*{Beta}&\multirow{2}*{(0.5, 0.999)}&{Adam coefficients used for computing
            }\\
      & \multirow{-2}*{Beta} &\multirow{-2}*{(0.5, 0.999)}&{gradients and their squares}\\
      &  Max epoch &30&-\\
      &  Batch size &256&-\\
         \midrule
    \multirow{5} *{Finetuning }  &     Optimizer & Adam &-\\
        & Learning rate &0.00005& Cosine scheduling\\
        & \multirow{2}*{Beta }&\multirow{2}*{(0.9, 0.999)}&{Adam coefficients used for computing
            }\\
       &\multirow{-2}*{Beta} &\multirow{-2}*{(0.9, 0.999)}&{gradients and their squares}\\
    &    Max epoch &200&-\\
     &   Batch size &32&-\\
            \bottomrule
    \end{tabular}
\end{table}

\begin{table}[H]
    \centering
    \caption{Hyperparameter settings for Method 1 with SSL-QA}
    \label{tab:Baseline hyper-paramter5}
    \begin{tabular}{llcl}
    \toprule
        & Name & Value & Description \\
         \midrule
          \multirow{5} *{Pretraining }   &Optimizer & Adam &-\\
        & Learning rate &0.0001& Cosine scheduling\\
         &\multirow{2}*{Beta}&\multirow{2}*{(0.5, 0.999)}&{Adam coefficients used for computing
            }\\
      & \multirow{-2}*{Beta} &\multirow{-2}*{(0.5, 0.999)}&{gradients and their squares}\\
      &  Max epoch &30&-\\
      &  Batch size &256&-\\
         \midrule
 \multirow{5} *{Finetuning }  &        Optimizer & Adam &-\\
      &   Learning rate &0.00005& Cosine scheduling\\
       &  \multirow{2}*{Beta }&\multirow{2}*{(0.9, 0.999)}&{Adam coefficients used for computing
            }\\
       &\multirow{-2}*{Beta} &\multirow{-2}*{(0.9, 0.999)}&{gradients and their squares}\\
    &    Max epoch &200&-\\
     &   Batch size &32&-\\
            \bottomrule
    \end{tabular}
\end{table}

\begin{table}[H]
    \centering
    \caption{Hyperparameter settings for Method 1 with joint pretraining}
    \label{tab:Baseline hyper-paramter5}
    \begin{tabular}{llcl}
    \toprule
 & Name & Value & Description \\
          \midrule
            \multirow{5} *{Pretraining }   &Optimizer & Adam &-\\
        & Learning rate &0.0001& Cosine scheduling\\
         &\multirow{2}*{Beta}&\multirow{2}*{(0.5, 0.999)}&{Adam coefficients used for computing
            }\\
      & \multirow{-2}*{Beta} &\multirow{-2}*{(0.5, 0.999)}&{gradients and their squares}\\
      &  Max epoch &30&-\\
      &  Batch size &256&-\\
         \midrule
    \multirow{7} *{Finetuning }  &        Optimizer & Adam &-\\
 & Optimizer for ignoring variables& Adam &-\\
      &   Learning rate &0.00005& Cosine scheduling\\
        &  Learning rate for ignoring variables&0.1&  Weight decay with a rate of 3e-4\\
       &  \multirow{2}*{Beta }&\multirow{2}*{(0.9, 0.999)}&{Adam coefficients used for computing
            }\\
       &\multirow{-2}*{Beta} &\multirow{-2}*{(0.9, 0.999)}&{gradients and their squares}\\
    &    Max epoch &200&-\\
     &   Batch size &32&-\\
            \bottomrule
    \end{tabular}
\end{table}

\begin{table}[H]
    \centering
    \caption{Hyperparameter settings for Method 1 with joint pretraining+ignoring}
    \label{tab:Baseline hyper-paramter5}
    \begin{tabular}{llcl}
    \toprule
        & Name & Value & Description \\
         \midrule
          \multirow{5} *{Pretraining }   &Optimizer & Adam &-\\
        & Learning rate &0.0001& Cosine scheduling\\
         &\multirow{2}*{Beta}&\multirow{2}*{(0.5, 0.999)}&{Adam coefficients used for computing
            }\\
      & \multirow{-2}*{Beta} &\multirow{-2}*{(0.5, 0.999)}&{gradients and their squares}\\
      &  Max epoch &30&-\\
      &  Batch size &256&-\\
         \midrule
 \multirow{7} *{Finetuning }  &        Optimizer & Adam &-\\
 & Optimizer for ignoring variables& Adam &-\\
      &   Learning rate &0.00005& Cosine scheduling\\
        &  Learning rate for ignoring variables&0.1&  Weight decay with a rate of 3e-4\\
       &  \multirow{2}*{Beta }&\multirow{2}*{(0.9, 0.999)}&{Adam coefficients used for computing
            }\\
       &\multirow{-2}*{Beta}
       &\multirow{-2}*{(0.9, 0.999)}&{gradients and their squares}\\
        &  \multirow{2}*{Beta for ignoring variables }&\multirow{2}*{(0.5, 0.999)}&{Adam coefficients used for computing
            }\\
       &\multirow{-2}*{Beta for ignoring variables} &\multirow{-2}*{(0.5, 0.999)}&{gradients and their squares}\\
    &    Max epoch &120&-\\
     &   Batch size &16&-\\
            \bottomrule
    \end{tabular}
\end{table}

\begin{table}[H]
    \centering
    \caption{Hyperparameter settings for Method 2}
    \label{tab:Baseline hyper-paramter6}
    \begin{tabular}{lcl}
    \toprule
         Name & Value & Description \\
         \midrule
         Optimizer & Adam &-\\
         Learning rate &0.01& Cosine scheduling\\
         \multirow{2}*{Beta }&\multirow{2}*{(0.5, 0.999)}&{Adam coefficients used for computing
            }\\
       \multirow{-2}*{Beta} &\multirow{-2}*{(0.5, 0.999)}&{gradients and their squares}\\
        Max epoch &200&-\\
        Batch size &256&-\\
            \bottomrule
    \end{tabular}
\end{table}

\begin{table}[H]
    \centering
    \caption{Hyperparameter settings for Method 2 with ignoring}
    \label{tab:Baseline hyper-paramter7}
    \begin{tabular}{lcl}
    \toprule
         Name & Value & Description \\
         \midrule
         Optimizer & Adam &-\\
         Optimizer for ignoring variables& Adam &-\\
         Learning rate &0.1& Cosine scheduling\\
         Learning rate for ignoring variables&0.01&  Weight decay with a rate of 3e-4\\
         \multirow{2}*{Beta }&\multirow{2}*{(0.9, 0.999)}&{Adam coefficients used for computing
            }\\
       \multirow{-2}*{Beta}
       &\multirow{-2}*{(0.9, 0.999)}&{gradients and their squares}\\
        \multirow{2}*{Beta for ignoring variables }&\multirow{2}*{(0.5, 0.999)}&{Adam coefficients used for computing
            }\\
       \multirow{-2}*{Beta for ignoring variables} &\multirow{-2}*{(0.5, 0.999)}&{gradients and their squares}\\
        Max epoch &180&-\\
        Batch size &64&-\\
            \bottomrule
    \end{tabular}
\end{table}

\begin{table}[H]
    \centering
    \caption{Hyperparameter settings for Method 2 with CMSSL-IQ}
    \label{tab:Baseline hyper-paramter8}
    \begin{tabular}{llcl}
        \toprule
        & Name & Value & Description \\
         \midrule
        \multirow{5} *{Pretraining }   &Optimizer & Adam &-\\
        & Learning rate &0.1& Cosine scheduling\\
         &\multirow{2}*{Beta}&\multirow{2}*{(0.5, 0.999)}&{Adam coefficients used for computing
            }\\
      & \multirow{-2}*{Beta} &\multirow{-2}*{(0.5, 0.999)}&{gradients and their squares}\\
      &  Max epoch &120&-\\
      &  Batch size &256&-\\
         \midrule
     \multirow{5} *{Finetuning }    & Optimizer & Adam &-\\
     &    Learning rate &0.01& Cosine scheduling\\
     &    \multirow{2}*{Beta }&\multirow{2}*{(0.5, 0.999)}&{Adam coefficients used for computing
            }\\
     &  \multirow{-2}*{Beta} &\multirow{-2}*{(0.5, 0.999)}&{gradients and their squares}\\
    &    Max epoch &200&-\\
    &    Batch size &256&-\\
            \bottomrule
    \end{tabular}
\end{table}

\begin{table}[H]
    \centering
    \caption{Hyperparameter settings for Method 2 with CMSSL-IA}
    \label{tab:Baseline hyper-paramter9}
    \begin{tabular}{llcl}
        \toprule
        & Name & Value & Description \\
         \midrule
        \multirow{5} *{Pretraining }   &Optimizer & Adam &-\\
        & Learning rate &0.1& Cosine scheduling\\
         &\multirow{2}*{Beta}&\multirow{2}*{(0.5, 0.999)}&{Adam coefficients used for computing
            }\\
      & \multirow{-2}*{Beta} &\multirow{-2}*{(0.5, 0.999)}&{gradients and their squares}\\
      &  Max epoch &120&-\\
      &  Batch size &256&-\\
         \midrule
     \multirow{5} *{Finetuning }    & Optimizer & Adam &-\\
     &    Learning rate &0.01& Cosine scheduling\\
     &    \multirow{2}*{Beta }&\multirow{2}*{(0.5, 0.999)}&{Adam coefficients used for computing
            }\\
     &  \multirow{-2}*{Beta} &\multirow{-2}*{(0.5, 0.999)}&{gradients and their squares}\\
    &    Max epoch &200&-\\
    &    Batch size &256&-\\
            \bottomrule
    \end{tabular}
\end{table}

\begin{table}[H]
    \centering
    \caption{Hyperparameter settings for Method 2 with SSL-QA}
    \label{tab:Baseline hyper-paramter10}
   \begin{tabular}{llcl}
        \toprule
        & Name & Value & Description \\
         \midrule
        \multirow{5} *{Pretraining }   &Optimizer & Adam &-\\
        & Learning rate &0.1& Cosine scheduling\\
         &\multirow{2}*{Beta}&\multirow{2}*{(0.5, 0.999)}&{Adam coefficients used for computing
            }\\
      & \multirow{-2}*{Beta} &\multirow{-2}*{(0.5, 0.999)}&{gradients and their squares}\\
      &  Max epoch &120&-\\
      &  Batch size &256&-\\
         \midrule
     \multirow{5} *{Finetuning }    & Optimizer & Adam &-\\
     &    Learning rate &0.01& Cosine scheduling\\
     &    \multirow{2}*{Beta }&\multirow{2}*{(0.5, 0.999)}&{Adam coefficients used for computing
            }\\
     &  \multirow{-2}*{Beta} &\multirow{-2}*{(0.5, 0.999)}&{gradients and their squares}\\
    &    Max epoch &200&-\\
    &    Batch size &256&-\\
            \bottomrule
    \end{tabular}
\end{table}

\begin{table}[H]
    \centering
    \caption{Hyperparameter settings for Method 2 with joint pretraining}
    \label{tab:Baseline hyper-paramter5}
    \begin{tabular}{llcl}
    \toprule
     & Name & Value & Description \\
          \midrule
            \multirow{5} *{Pretraining }   &Optimizer & Adam &-\\
        & Learning rate &0.0001& Cosine scheduling\\
         &\multirow{2}*{Beta}&\multirow{2}*{(0.5, 0.999)}&{Adam coefficients used for computing
            }\\
      & \multirow{-2}*{Beta} &\multirow{-2}*{(0.5, 0.999)}&{gradients and their squares}\\
      &  Max epoch &30&-\\
      &  Batch size &256&-\\
         \midrule
    \multirow{7} *{Finetuning }  &        Optimizer & Adam &-\\
      &   Learning rate &0.00005& Cosine scheduling\\
       &  \multirow{2}*{Beta }&\multirow{2}*{(0.9, 0.999)}&{Adam coefficients used for computing
            }\\
       &\multirow{-2}*{Beta}
       &\multirow{-2}*{(0.9, 0.999)}&{gradients and their squares}\\
    &    Max epoch &200&-\\
     &   Batch size &256&-\\
            \bottomrule
    \end{tabular}
\end{table}

\begin{table}[H]
    \centering
    \caption{Hyperparameter settings for Method 2 with joint pretraining+ignoring}
    \label{tab:Baseline hyper-paramter5}
    \begin{tabular}{llcl}
    \toprule
        & Name & Value & Description \\
         \midrule
          \multirow{5} *{Pretraining }   &Optimizer & Adam &-\\
        & Learning rate &0.0001& Cosine scheduling\\
         &\multirow{2}*{Beta}&\multirow{2}*{(0.5, 0.999)}&{Adam coefficients used for computing
            }\\
      & \multirow{-2}*{Beta} &\multirow{-2}*{(0.5, 0.999)}&{gradients and their squares}\\
      &  Max epoch &120&-\\
      &  Batch size &256&-\\
         \midrule
 \multirow{7} *{Finetuning }  &        Optimizer & Adam &-\\
 & Optimizer for ignoring variables& Adam &-\\
      &   Learning rate &0.00005& Cosine scheduling\\
        &  Learning rate for ignoring variables&0.1&  Weight decay with a rate of 3e-4\\
       &  \multirow{2}*{Beta }&\multirow{2}*{(0.9, 0.999)}&{Adam coefficients used for computing
            }\\
       &\multirow{-2}*{Beta} &\multirow{-2}*{(0.9, 0.999)}&{gradients and their squares}\\
       &  \multirow{2}*{Beta for ignoring variables }&\multirow{2}*{(0.5, 0.999)}&{Adam coefficients used for computing
            }\\
       &\multirow{-2}*{Beta for ignoring variables} &\multirow{-2}*{(0.5, 0.999)}&{gradients and their squares}\\
    &    Max epoch &180&-\\
     &   Batch size &64&-\\
            \bottomrule
    \end{tabular}
\end{table}

\section{Files containing trained models}
Files containing  trained models are listed in Table~\ref{table:pt}. 
\begin{table}[H]
    \centering
    \caption{Trained models for different methods}
    \begin{tabular}{ll}
    \toprule
         Name & Links  \\
          \midrule
         Faster RCNN pretrained on BCCD&faster\_rcnn\_res101\_bccd.pth$^2$\\
         \hline
Method 1 without image& Method 1/m1\_no\_image.pth$^2$\\
Method 1 & Method 1/m1.pth$^2$\\
Method 1 with ignoring
& Method 1/m1\_ignoring.pth$^2$\\
Method 1 with CMSSL-IQ
& Method 1/m1\_IQ.pth$^2$\\
Method 1 with CMSSL-IA
& Method 1/m1\_IA.pth$^2$\\
Method 1 with SSL-QA
& Method 1/m1\_QA.pth$^2$\\
Method 1 with joint pretraining
& Method 1/m1\_joint.pth$^2$ \\
Method 1 with joint pretraining+ignoring
& Method 1/m1\_joint\_ignoring.pth$^2$ \\
\hline
Method 2 without image& Method 2/m2\_no\_image.pth$^2$\\
Method 2 & Method 2/m2.pth$^2$\\
Method 2 with ignoring
& Method 2/m2\_ignoring.pth$^2$\\
Method 2 with CMSSL-IQ
& Method 2/m2\_IQ.pth$^2$\\
Method 2 with CMSSL-IA
& Method 2/m2\_IA.pth$^2$\\
Method 2 with SSL-QA
& Method 2/m2\_QA.pth$^2$\\
Method 2 with joint pretraining
& Method 2/m2\_joint.pth$^2$ \\
Method 2 with joint pretraining+ignoring
& Method 2/m2\_joint\_ignoring.pth$^2$ \\
   \bottomrule
\vspace{-0.3cm}
    \end{tabular}
\footnotesize{$^2$\url{https://drive.google.com/drive/folders/1tTibOSHNcmMc9rXD7xIwWV4F41ee3A0w?usp=sharing}}
\label{table:pt}
\end{table}

\section{Training time}
Table~\ref{tab:trainingtime} shows the training time for each experiment using GTX 1080ti GPUs with the hyper-parameter settings described above.
\begin{table}[H]
    \centering
        \caption{Training time for different experiments}
    \label{tab:trainingtime}
    \begin{tabular}{l|c}
    \toprule
        Name & Training time  \\
        \midrule
       Method 1 without image& 5 hour\\
Method 1 & 13 hour\\
Method 1 with ignoring
& 34 hour \\
Method 1 with CMSSL-IQ
& 21 hour\\
Method 1 with CMSSL-IA
& 24 hour\\
Method 1 with SSL-QA
& 23 hour\\
Method 1 with joint pretraining
& 36 hour \\
Method 1 with joint pretraining+ignoring
&  57 hour\\
\hline
Method 2 without image  & 1.5 hour\\
Method 2 
& 3 hour \\
Method 2 with ignoring
& 17 hour\\
Method 2 with CMSSL-IQ 
& 4 hour\\
Method 2 with CMSSL-IA
& 4 hour\\
Method 2 with SSL-QA
& 3 hour\\
Method 2 with joint pretraining
&  11 hour\\
Method 2 with joint pretraining+ignoring
&  25 hour\\
    \bottomrule
    \end{tabular}
\end{table}
\section{Parameter numbers}
Table~\ref{tab:Parameter number} shows the parameter numbers for different models.
\begin{table}[H]
    \centering
        \caption{Parameter number of different models}
    \label{tab:Parameter number}
    \begin{tabular}{l|c}
    \toprule
         Model  & Parameter numbers  \\
         \midrule
       Method 1 without image&120 million\\
Method 1 & 238 million\\
Method 1 with ignoring
& 238 million\\
Method 1 with CMSSL-IQ
& 257 million\\
Method 1 with CMSSL-IA
& 257 million \\
Method 1 with SSL-QA
& 257 million \\
Method 1 with joint pretraining
& 257 million \\
Method 1 with joint pretraining+ignoring
& 258 million \\
\hline
Method 2 without image  &137 million \\
Method 2 
& 84 million \\
Method 2 with ignoring
& 84 million\\
Method 2 with CMSSL-IQ 
& 93 million\\
Method 2 with CMSSL-IA
& 93 million\\
Method 2 with SSL-QA
& 137 million\\
Method 2 with joint pretraining
& 138 million \\
Method 2 with joint pretraining+ignoring
& 138 million \\
    \bottomrule
    \end{tabular}
\end{table}

\bibliographystyle{iclr2021_conference}